\documentclass{article}

\usepackage{arxiv}

\usepackage{amsthm}
\usepackage{mathrsfs,amsmath}
\usepackage{newtxtext,newtxmath}
\usepackage{arydshln}
\usepackage{float}
\usepackage{subfig}
\usepackage{xcolor}

\usepackage[utf8]{inputenc} 
\usepackage[T1]{fontenc}  
\usepackage{booktabs}  
\usepackage{amsfonts}
\usepackage{nicefrac} 
\usepackage{microtype}
\usepackage{cleveref} 
\usepackage{lipsum} 
\usepackage{graphicx}
\usepackage[numbers]{natbib}
\usepackage{doi}

\title{Feather: An Elegant Solution to Effective DNN Sparsification}

\date{}
 
\author{ 
	Athanasios Glentis Georgoulakis\\
	School of ECE, NTUA, Athens, Greece \\
	\texttt{athglentis@gmail.com} \\
	\And
	George Retsinas \\
	School of ECE, NTUA, Athens, Greece \\
	\texttt{gretsinas@central.ntua.gr} \\
	\And
	Petros Maragos \\
	School of ECE, NTUA, Athens, Greece \\
	\texttt{maragos@cs.ntua.gr} \\
}

\renewcommand{\headeright}{}
\renewcommand{\shorttitle}{Feather: An Elegant Solution to Effective DNN Sparsification}

\usepackage{xspace}

\makeatletter
\DeclareRobustCommand\onedot{\futurelet\@let@token\@onedot}
\def\@onedot{\ifx\@let@token.\else.\null\fi\xspace}

\newcommand{\vs}{\emph{vs}\onedot} 
\def\eg{\emph{e.g}\onedot}

\def\ie{\emph{i.e}\onedot}
\def\wrt{{w.r.t}\onedot}
\def\bs#1{\boldsymbol{#1}}

\makeatother

\newcommand{\res}[2]{#1 ${\scriptstyle \pm #2}$ }
\newcommand{\bres}[2]{\textbf{#1} ${\scriptstyle \pm #2}$ }

\begin{document}
\maketitle

\begin{abstract}
	Neural Network pruning is an increasingly popular way for producing compact and efficient models, suitable for resource-limited environments, while preserving high performance. While the pruning can be performed using a multi-cycle training and fine-tuning process, the recent trend is to encompass the sparsification process during the standard course of training. To this end, we introduce Feather, an efficient sparse training module utilizing the powerful Straight-Through Estimator as its core, coupled with a new thresholding operator and a gradient scaling technique, enabling robust, out-of-the-box sparsification performance. Feather’s effectiveness and adaptability is demonstrated using various architectures on the CIFAR dataset, while on ImageNet it achieves state-of-the-art Top-1 validation accuracy using the ResNet-50 architecture, surpassing existing methods, including more complex and computationally heavy ones, by a considerable margin. Code is publicly available at {\url{https://github.com/athglentis/feather}}.

\end{abstract}


\section{Introduction}

Machine learning applications have benefited tremendously in the past decade from the use of Deep Neural Networks (DNNs), especially in the field of computer vision \cite{lecun2015deep}. While DNNs are more than capable at achieving state-of-the-art (SoA) results, they rely upon having very large numbers of trained parameters and thus require huge amounts of computational and memory resources \cite{krizhevsky2012imagenet,  taigman2014deepface, he2016deep}. Those requirements often prohibit the use of powerful DNNs in mobile and resource-limited devices, resulting in a gap between the achieved SoA results and the ones coming from the deployed technology. 
Recognizing that DNNs are heavily over-parametrized \cite{sze2017efficient}, network pruning (\ie the process of removing network parameters) has been studied extensively in recent years as a way of drastically reducing the model's size and computational footprint \cite{cheng2018model, deng2020model,liang2021pruning, hoefler2021sparsity}. 

Broadly the pruning methodology can be categorized in terms of (a) the granularity of the sparsified elements, (b) the pruning criterion and (c) the timeframe the sparsity is induced.
Based on granularity methods are divided between structured or unstructured. Structured methods remove groups of parameters such as kernels and filters \cite{li2016pruning, molchanov2016pruning}, attaining sparsity that is more easily utilized by commodity hardware for computational gains, while unstructured freely remove individual parameters, generally leading to more accurate sparse models \cite{han2015deep, han2015learning}.
As pruning criterion, a large number of methods associate parameter importance with large magnitude values and prune elements with magnitudes smaller than some threshold value \cite{han2015deep, han2015learning, zhu2017prune}. This simple criterion, judging a weight's importance based on its magnitude, is efficient to compute and is found to be effective in the pruning literature, achieving high sparsity ratios with minimal performance loss \cite{han2015learning, gale2019state}. An alternative way involves estimating how pruning an element will affect the training loss, however requires calculating Hessian information which is computationally expensive \cite{hassibi1993optimal,dong2017learning, aghasi2020fast}.
Finally, considering at what stage of the training process the pruning takes place, the different algorithms are usually divided into dense-to-sparse and sparse-to-sparse training \cite{kusupati2020soft}. The former, starting with a dense network produce the requested sparse version at the end of the pruning procedure \cite{han2015learning,kusupati2020soft,zhu2017prune}. Some methods of this type require multiple pruning and retraining cycles \cite{renda2020comparing} and thus prolong the required training time. Sparse-to-sparse methodologies aim to reduce training times even further by starting training an already sparse network\cite{mocanu2018scalable, dettmers2019sparse, mocanu2021sparse} although the achieved performances generally fall behind of those using dense-to-sparse training methods \cite{kusupati2020soft}.\par 
 
More recently, a number of works \cite{retsinas2021online, jayakumar2020top, tai2022spartan, vanderschueren2023straight} which are built around the concept of the Straight-Through Estimator (STE) \cite{bengio2013estimating} have demonstrated that SoA results can be achieved by pruning DNNs along the normal course of training. Sparse training with the STE is performed by computing the forward pass using the thresholded (pruned) version of the weights while updating the dense weights during the backward pass, treating the thresholding function (that performs the pruning) as the identity. 

In our work, we focus on improving sparse training with the STE by addressing a number of overlooked shortcomings of the method, eventually introducing a novel pruning module. We propose (i) 
a new thresholding function used for magnitude pruning and (ii) a straightforward way to control gradient flow of the pruned weights. 
More specifically, instead of using hard or soft thresholding \cite{donoho1995noising}, which previous STE based methods mostly use, we propose a family of thresholding functions that lie in between the aforementioned two and combine their advantages, namely reduced bias between the thresholded weights and their dense counterparts and a smooth transition region near the threshold. 
Complementary to the proposed thresholding approach, we suggest scaling the gradients attributed to pruned parameters by a parameter $\theta \in (0,1)$, aiming on improving the stability of the pruning mask, a factor that we find to be crucial when targeting very high sparsity ratios.

We demonstrate the effectiveness of our sparse training approach when applied to magnitude based unstructured pruning frameworks, which reach a user-specified sparsity ratio by incrementally pruning the network during training. In more detail, we perform extensive experiments on both CIFAR \cite{krizhevsky2009learning} and ImageNet \cite{deng2009imagenet} datasets using a very simplistic global thresholding pruning procedure as a backbone and additionally with a recently proposed layer-wise pruning framework \cite{retsinas2021online}. Our sparse training approach surpasses the (generally more computationally expensive) current SoA unstructured pruning algorithms, significantly improving the previously achieved generalization accuracies of the resulting sparse models.  

Overall, the contributions of our paper can be summarized as follows: 
\begin{itemize}
    \item 
    We introduce Feather, a versatile sparse training module that can be used to efficiently and effectively prune neural networks up to extreme sparsity   levels. 
    The proposed module was evaluated on different magnitude pruning backbones and achieved consistent improvements over the prior state-of-the-art.
    \item We highlight the importance of a well-crafted thresholding function that finds a fine balance between the two standard ones, namely hard and soft operators.
    \item We highlight the correlation between scaling the gradients of the pruned weights and the targeted sparsity ratio: high sparsity targets should be accompanied with lower scaling values to provide high performing pruned models.
\end{itemize}

\section{Proposed Method}

In this work, we developed a novel pruning module, dubbed as Feather. The name symbolizes its lightweight nature, the elegance with which it achieves sparsity, as well as the lightness of the resulting pruned networks. It can be utilized in various  magnitude pruning schemes,  including strategies where pruning can be performed globally or in a layer-wise fashion.
The setting of interest is magnitude pruning, where a weight is kept only if its magnitude surpasses a threshold value $T$, 
and a sparse training process is followed (i.e., perform the pruning procedure along the standard course of training), as done in most of the SoA systems. 
In essence, we capitalize on the effectiveness of modern STE-based approaches, carefully addressing potential issues. 
Implementation-wise, the proposed module is applied at each layer, replacing a typical pruning operation, affecting both the forward and the backward step. 

In what follows, we first provide the necessary prerequisites for sparse training with STE and then we describe the proposed module, emphasizing on the proposed modifications on both the forward and the backward steps and highlighting the underlying motivations.

\subsection{Preliminaries: Sparse Training}

In this analysis we examine how the pruning and the subsequent weight update is performed (during a training iteration) on a single layer under the STE framework.

First, let us consider a thresholding operator, the core of the magnitude pruning approaches, as a function $\mathcal{P}_{(T)}(x)$  that performs the pruning given a threshold value $T$. As a general rule this function takes zero values when $|x| \leq T$ and non-zero values otherwise. Typical instantiations of this function are the hard and the soft thresholding operators.

A first approach is to directly back-propagate through the thresholding operator. 
By doing so, the gradients belonging to the pruned weights will be zeroed thus excluding them from the update step. Regrettably, due to the resulting sparse gradient, this approach may lead to unwanted decay of immaturely trained weights and a slow exploration of the possible sparsity patterns \cite{jayakumar2020top, tai2022spartan}.

Lately, to circumvent the aforementioned issues, a number of pruning methods \cite{retsinas2021online, jayakumar2020top, tai2022spartan, vanderschueren2023straight} have achieved SoA results relying on the concept of a Straight-Through Estimator \cite{bengio2013estimating} training approach. In a nutshell, with the STE formulation, the weight update and thresholding equations are now decoupled into: 
\begin{eqnarray}
  \bs{\tilde{w}}_{k} & = & \mathcal{P}_{(T)}(\bs{w}_{k}) 
  \label{eq:ste1}\\
  \bs{w}_{k+1} & = & \bs{w}_k-\eta \cdot \nabla \mathcal{L}( \bs{\tilde{w}}_k), 
  \label{eq:ste2}
\end{eqnarray}
where $\bs{w}$ are the vectorized layer weights, $ \bs{\tilde{w}}$ the pruned weights after applying the thresholding operation,  
$\mathcal{L}(\bs{w})$ the loss function and $\eta$ the step size.
The reported expressions correspond to $k$-th iteration of a Gradient Descent formulation to highlight the effect of STE. The same procedure is trivially extended to any optimizer required.

The main idea is to consider the thresholding operator as the identity function during back-propagation and update both pruned and unpruned weights based on the gradients of the loss \wrt the sparse set of weights. During the forward pass however, only the sparse weights are used so that the network is trained under the sparsity constrain. The key benefit of sparse training with the STE is that it allows for pruned weights to become active again, if at some point during the training they get a large enough magnitude. This process therefore promotes the exploration of different sparsity patterns and is found to result to better performing sparse networks \cite{jayakumar2020top, tai2022spartan}.

\subsection{Proposed Sparse Training Module}

\begin{figure}[tp]
\begin{center}
\subfloat[]{\includegraphics[width=.52\linewidth]{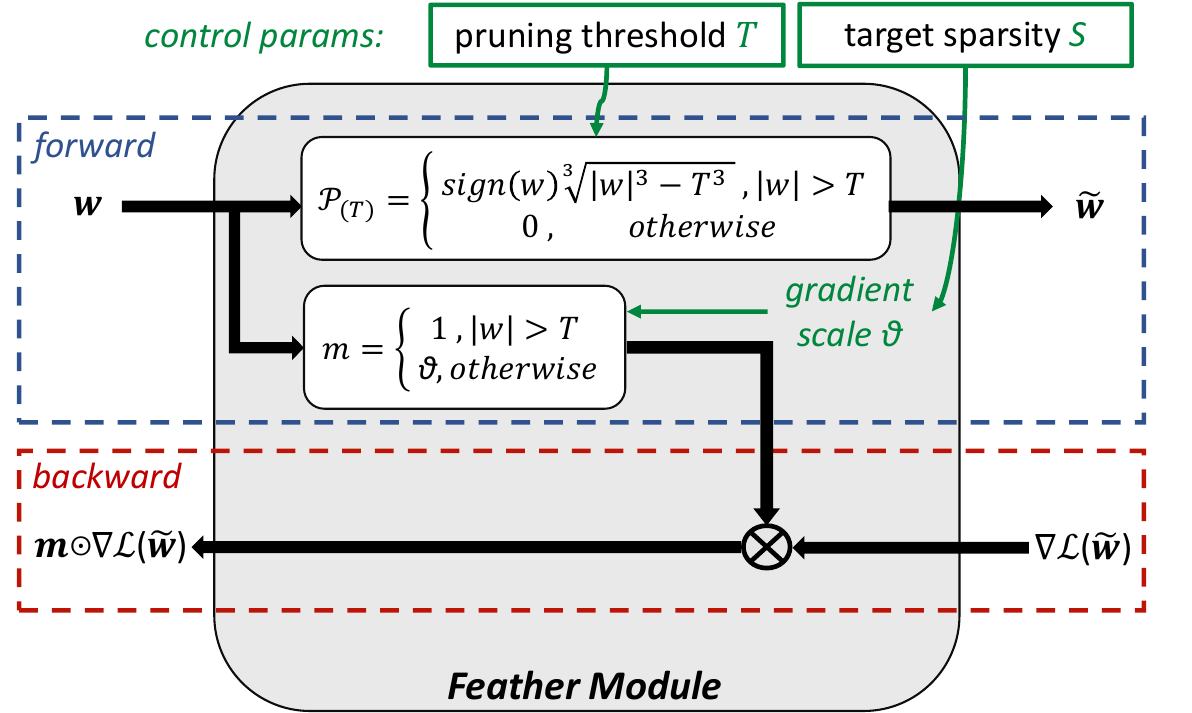}
\label{fig:block}
}
\subfloat[]{\includegraphics[width=.45\linewidth]{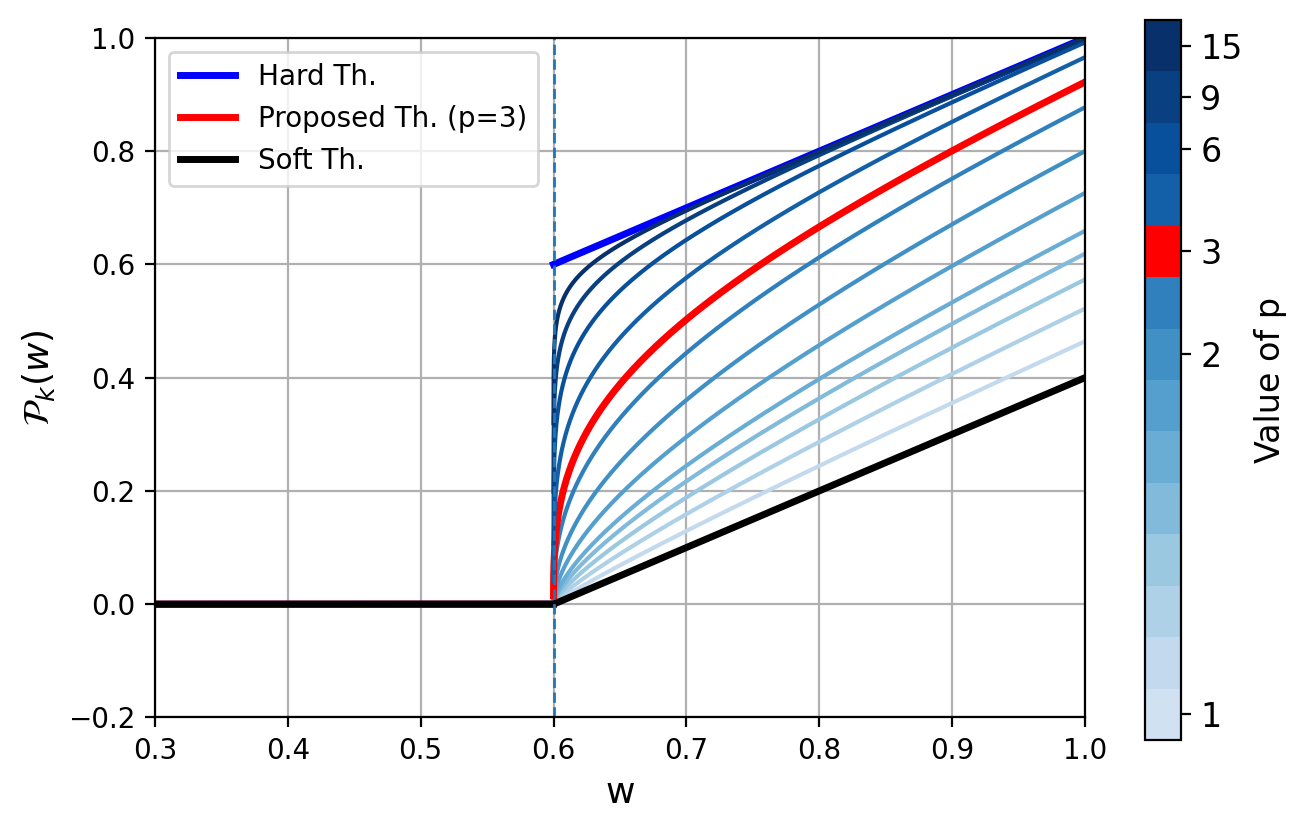}
\label{fig:thresholds}
}
\caption{(a) The proposed sparse training block, utilizing the new thresholding operator and the gradient scaling mask (b) the proposed family of thresholding operators for varying values of $p$. We adopt $p=3$, resulting to a fine balance between the two extremes, hard and soft thresholding respectively.}
\end{center}
\end{figure}

The aforementioned STE pipeline was a stimulus for the proposed module; 
our primary goal was to maintain the simplicity of the original idea, and thus we concentrated on enhancing two fundamental elements: the thresholding operator during the forward step and the gradient manipulation during the backpropagation step.
Overall the proposed enhancements aim to assist the training process by promoting convergence to well-performing yet highly sparse solutions.
The functionality of Feather, the proposed module, is summarized in Figure \ref{fig:block}, where the depicted components will be described in detail in what follows.

\subsubsection{STE Thresholding Operator}

Focusing on the thresholding operator used during the forward pass, the most straightforward way to define a magnitude pruning step is through the hard thresholding function, which is discontinuous at the threshold value. This discontinuity might result to training instabilities when weights pass from pruned to active states and vice versa, since the gradient received might not justify the value after thresholding. To addressing this matter, a recent method \cite{vanderschueren2023straight} suggested that the soft thresholding operator is preferable in the context of STE, while soft thresholding is a common choice in pruning approaches in general~\cite{kusupati2020soft}. Soft thresholding, although suppressing the discontinuity, induces a constant bias (equal to the threshold value) between the active weights and their dense counterparts, updated during the STE back-propagation phase. Note that by its nature the STE introduces an inconsistency between the forward and the backward pass, since during the former the sparse set of weights is used whereas during the later the computed gradients are used to update the dense weights.

Motivated by works in the area of sparse regression, aiming on finding thresholding operators that lie in between hard and soft thresholding \cite{liu2020between, hagiwara2022bridging}, we propose the following family of operators to be used with STE:
\begin{equation}
	\mathcal{P}_{(T)}(w) = 
\left\{
	\begin{array}{ll}
		\textrm{sign}(w) \cdot (|w|^p - T^p)^{1/p},  & \textrm{if } |w| > T \\
		0,  & \textrm{ otherwise} \\
	\end{array} \right.\\	
\end{equation}
The behavior of $\mathcal{P}_{(T)}$ is depicted in Figure \ref{fig:thresholds}, for varying values of the power parameter $p$. Intuitively, as $p$ grows we deviate from the soft ($p=1$) and approach the hard thresholding operator. Under that point of view, the proposed operator can be considered a generalization of the existing two.  Note that the proposed function, when avoiding extreme selections for $p$, tries to balance between the two aforementioned properties: continuity and bias. Based on that, we suggest a value of $p=3$ being a reasonable compromise that adequately addresses both issues. 
We provide further validating experiments in the Appendix.

In a final note, we want to emphasize that our proposed thresholding operator, apart from having an intuitive explanation for the resulting improved performance (as empirically evaluated in Section \ref{sec:experimental}), due to its simplicity it practically imposes no training overhead and can be used off-the-self in pruning frameworks in conjunction with the STE.

\subsubsection{STE and Gradient Scaling}
\label{sec:scaling}

The main motivation behind the STE is to allow gradient flow to pruned weights and thus enable the exploration of multiple sparsity patterns during sparse training. As stated before, this is achieved by treating the thresholding function as the identity during the back-propagation. However, in certain cases it might be beneficial to limit the variations of the mask and favor a more stable sparsity pattern. We find that a rather straightforward, yet intuitive way to control the mask's stability is to scale the gradients of the pruned weights by a constant value $\theta \in [0,1]$, effectively modifying the update step of Eq. \ref{eq:ste2} into:
\begin{equation}
\bs{w}_{k+1}  =  \bs{w}_k - \eta \cdot \bs{m}_k \odot \nabla \mathcal{L}( \bs{\tilde{w}}_k), 
\end{equation}
where $\bs{m}_k \in \{\theta, 1\}^N$ such that $m_{i,k} = 1$ if $w_{i,k} > T$ and  $m_{i,k} = \theta$ otherwise, with $\odot$ denoting element-wise product.

At the two extreme points, $\theta = 0$ and $\theta = 1$, the method aligns with the non-STE and the standard STE approaches. For in-between values of $\theta$, pruned weights continue to receive gradient updates but with scaled down magnitudes, and therefore are, to some extent, promoted to decay, resulting to a more stable (but not completely fixed) mask.

In our experiments (Section \ref{sec:ablations}) we find that scaling the gradients becomes beneficial when targeting very high sparsity ratios  (\eg $98\%$ and above) where, due to having very few active weights, too frequent mask variations appear to destabilize the network. For more conservative ratios, using $\theta < 1$ appears not to improve the results and even lead to a small decline in accuracy at some cases. 

To this end, based on experimental evidence, essentially relying on a profiling approach, we define an ``automatic'' way to set $\theta$. 
In more detail, we select $\theta = g(S)$ as a simple step function, where $\theta = 1$ if the final target sparsity $S$ is under $95\%$ and $\theta = 0.5$ when $S \geq 95\%$. The value of $\theta$ is selected at the beginning of training and remains constant throughout the training process.
Interestingly, no performance gains were observed by adopting more complex scheduling tactics (e.g. having a gradual smooth transition over $\theta$ from 1 to a lower value) during training. Note that ideally a more complex function could be defined given an exhaustive profiling procedure, but we considered such ideas out of scope for this work.

\section{Application on Pruning Frameworks}
\label{sec:frameworks}

Our proposed sparse training module is versatile and not restricted to a particular pruning framework. We will demonstrate its effectiveness using two distinct frameworks, both a global and a layer-wise magnitude pruning backbone, as described bellow.

\hfill \break
\noindent \textbf{Global Magnitude Pruning:} Most pruning methods that utilize the STE (or some variant of it) use  global thresholding, in the sense that a single threshold $T$ is selected for all layers, computed by sorting all weights, in order to prune the network up-to a specified sparsity ratio \cite{jayakumar2020top, vanderschueren2023straight, tai2022spartan}. Common practice is to incrementally increase the requested ratio throughout the training process, thus giving the network time to adjust to different sparsity levels \cite{zhu2017prune,kusupati2020soft}. A popular sparsity schedule was proposed in \cite{zhu2017prune}, where the network is trained densely for a small number of warm up epochs followed by a cubical increase of the sparsity ratio, until it reaches the final target ratio, which is then kept constant for the rest of the training epochs. For simplicity and since the exact sparsity schedule is not the focus of this work, we reach the final target ratio at $50\%$ of the epochs, without having a densely training warm up phase which we found not to have an effect to the final performance. 

\hfill \break
\noindent \textbf{Layer-wise Magnitude Pruning:} 
To further reveal the efficacy of Feather, we adopted a considerably different pruning framework and pair it with the proposed module.
We considered the Adaptive Sparsity Loss framework (ASL)~\cite{retsinas2021online}, which tackles the magnitude pruning problem in an explicit layer-wise formulation. Specifically, ASL considers the per-layer pruning thresholds as trainable parameters which are included in an additional loss term (Sparsity Loss), constructed based on assumptions for the per-layer distributions. The thresholds are consequently learned during training by minimizing the extra loss term, resulting into a learned non-uniform per-layer sparsity. The overall sparsity of the DNN is constrained, via the loss, to a specific target sparsity. 
It should be noted that STE is also a core element of this approach. In this work, we used an enhanced version, referred to as ASL+, where we addressed existing shortcomings of the original ASL, including the addition of a sparsity scheduler, similar to the one defined in the global pruning counterpart, an element missing from \cite{retsinas2021online}, as well as a sorting-based correction step of the per-layer threshold to compensate for the approximate nature of the distribution-motivated sparsity loss. The alluring aspect of ASL is the ability to further define more complex sparsity-related goals through appropriately crafted sparsity losses, but such extension is outside the scope of this work.

\section{Experimental Evaluation}
\label{sec:experimental}

The present section provides the experimental validation of the proposed method's effectiveness. First, we perform ablation studies to showcase the efficacy of each element of our approach. Then, we provide comparisons with relevant baselines and SoA unstructured pruning algorithms. 
Specifically, we experiment with modern compact architectures ResNet-20 \cite{he2016deep}, MobileNetV1 \cite{howard2017mobilenets} and DenseNet40-24 \cite{huang2017densely} on CIFAR-100\footnote{The channels of ResNet-20 are doubled for the experiments on CIFAR, as also done in \cite{zhou2021effective, wang2020picking}.}  \cite{krizhevsky2009learning}.
Furthermore, we provide large-scale experiments on the ImageNet \cite{deng2009imagenet} dataset using the ResNet50 \cite{he2016deep} architecture in order to further verify the generalization abilities of our method and its performance edge over the current SoA. The reported results are obtained using SGD optimizer along with a Cosine Annealing scheduler, while the training hyperparameters are the typically used in literature for such settings (described in detail in the Appendix).

\subsection{Ablation Studies}
\label{sec:ablations}

The ablation studies were performed on the CIFAR-100 dataset, using the global magnitude pruning framework as the backbone to the Feather pruning module. We note that similar behaviors were observed using the Feather in conjunction with the layer-wise ASL\texttt{+} framework. All figure data points represent averages of 3 runs and the corresponding standard deviations are shown as shaded regions.

\begin{figure}[h]
	\centering
  \includegraphics[width=1.0\linewidth]{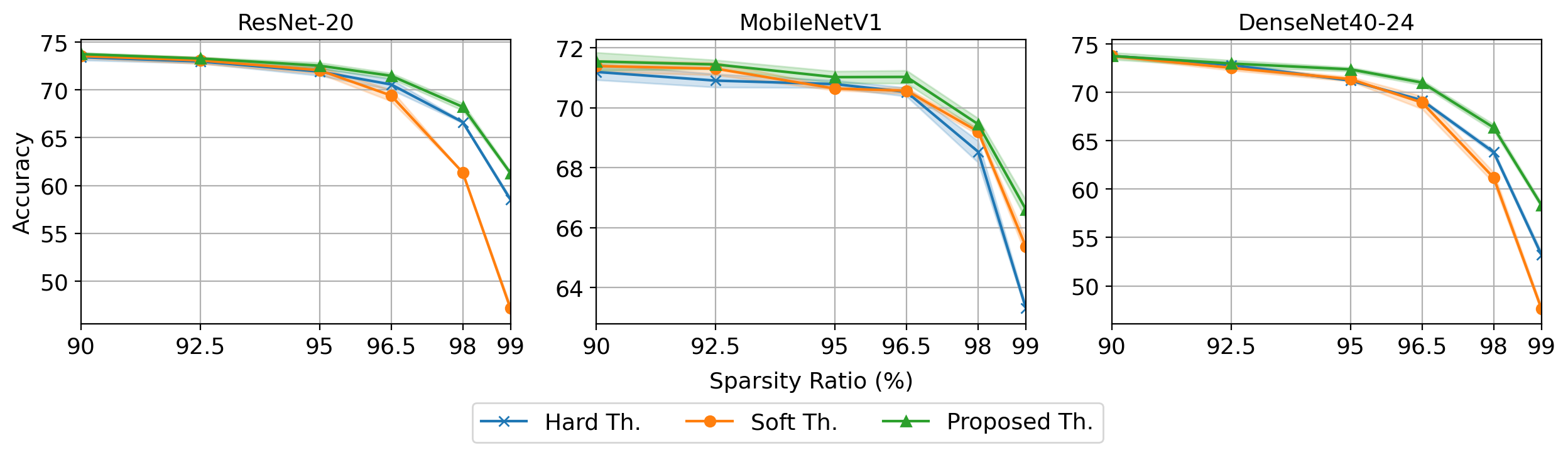}
\caption{Study of the effect of the thresholding operator on the final sparse model accuracy. The proposed threshold steadily outperforms the hard and soft operators.}
\label{fig:abl1}
\end{figure}

\noindent \textbf{Impact of Thresholding Operator:} 
\noindent The effectiveness of the proposed thresholding function (with $p=3$) is empirically evaluated and compared against that of the hard and soft threshold functions in Figure \ref{fig:abl1}, while no gradient scaling was considered for this experiment (we used typical STE). The proposed threshold enables the training of more accurate sparse networks, especially at high pruning ratios ($95\%$ and above). Notably, the hard thresholding approach achieves considerably low results with the MobileNetV1 network, while the soft threshold with the other two. 
\textit{Overall, our approach consistently leads to better trained networks, regardless the architecture and the sparsity ratio, supporting our claims of effectively combining the strengths of both soft and hard thresholding.}

\begin{figure}[H]
	\centering
  \includegraphics[width=.95\linewidth]{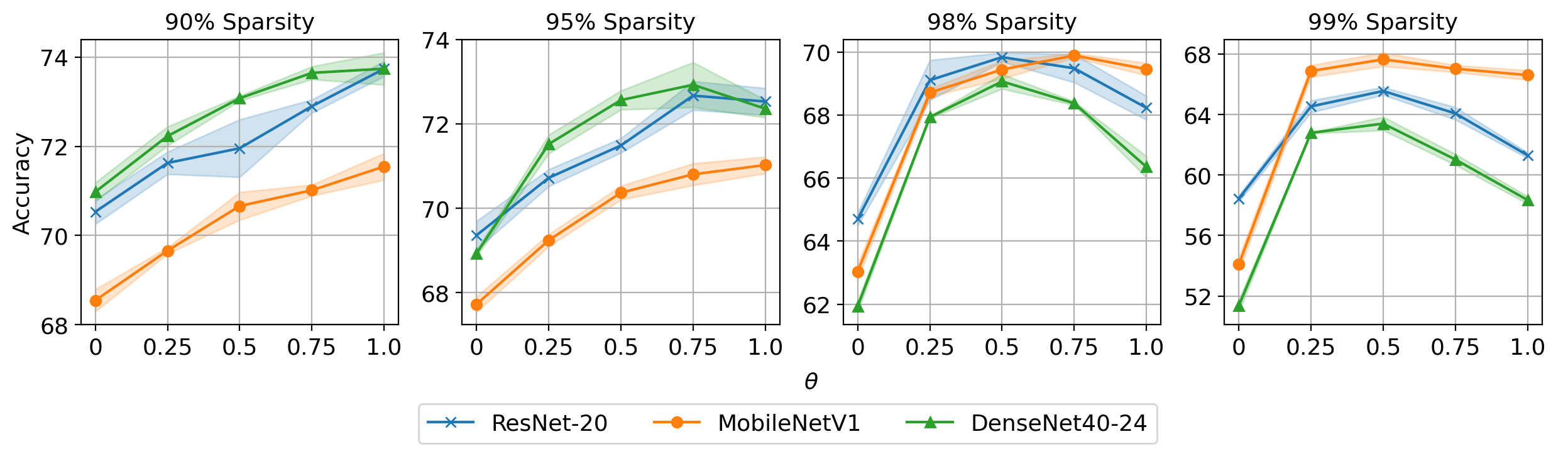}
\caption{Study of the effect of gradient scaling. Under conservative final sparsity, $\theta$ near unity is preferable, while when targeting high sparsity, models benefit from $\theta$ near the middle of its range.}
\label{fig:abl2}
\end{figure}

\noindent \textbf{Impact of Gradient Scaling:}   
In Figure \ref{fig:abl2} we investigate the relation between the value of the parameter $\theta \in [0,1]$, that scales the gradient of the pruned weights, and the final model accuracy when requesting different sparsity ratios. Note that the under-performing case of $\theta = 0$ is equivalent to a non-STE variant.
\textit{ Across all considered models, a clear trend can be seen; When targeting lower sparsity levels, best results are achieved with values of $\theta$ close to unity, whereas at more extreme sparsity ratios (such as $98\%$ and $99\%$) the optimal values of $\theta$ seem to shift towards the middle of its range.
}

This observed dependency is the motivation behind an automatic selection of $\theta$, as described in Section~\ref{sec:scaling}, where $\theta = 0.5$ for $S \geq 95\%$ and  $\theta = 1$ otherwise. Note that this selection just aligns with the reported trend and is not optimal for every case reported. 
Despite this being a very ``crude'' selection, it is very effective, as the forthcoming experimental evaluations hint.
Nonetheless the takeaway of this experiment is not a simple function of two modes, but bringing this relation to the spotlight.
Based on this observation, we pave the way towards more complex functions or, more interestingly, towards having different scales per layer, relying on the per-layer sparsity rather than the overall sparsity. The later idea, a possible future direction of practical value, simply states that under-pruned layers can be more flexible to sparsity pattern variations than the overly pruned ones.

\begin{figure}[H]
	\centering
  \includegraphics[width=1\linewidth]{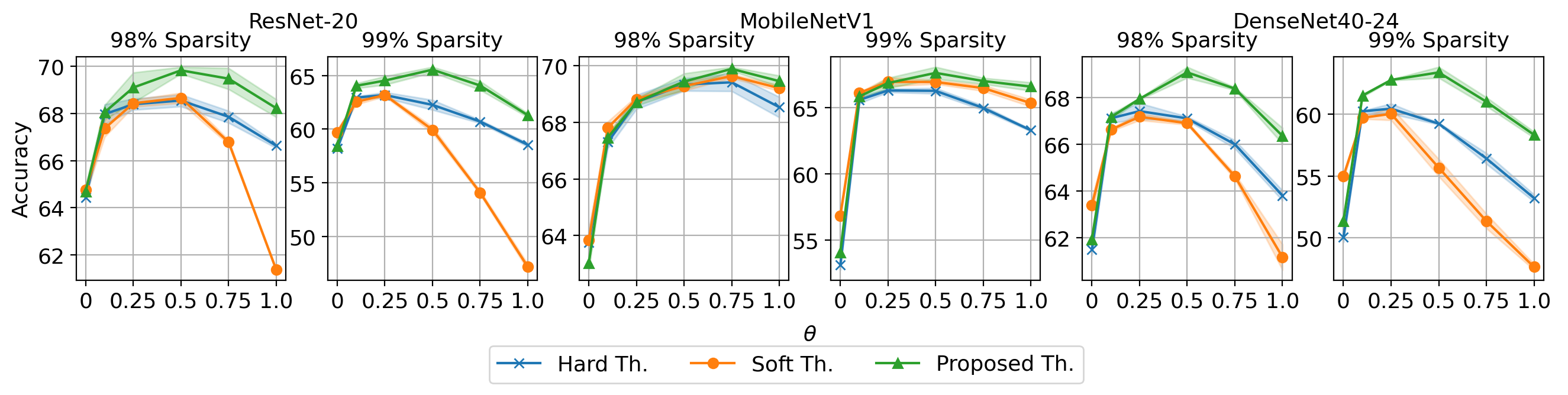}
\caption{Gradient scaling improves the final accuracy at high sparsity, regardless the thresholding operator, while maximum performance is achieved if combined with the proposed threshold.}
\label{fig:abl3}
\end{figure}

\noindent \textbf{Gradient Scaling under different Thresholding Functions:}
\noindent Finally, we compare the impact of gradient scaling using the three different types of thresholding operators. Figure \ref{fig:abl3}  shows that, regardless the choice of operator, using a scale $\theta \in (0,1)$ is beneficial to the final accuracy at the high sparsity regimes. Nevertheless, our threshold maintains the lead in performance compared to the two standard ones.

\subsection{Comparison to SoA}

\noindent \textbf{CIFAR-100:}
\noindent For these experiments, we directly compare our results with the ones of ST-3 \cite{vanderschueren2023straight} and Spartan \cite{tai2022spartan}, being the two most recent and best performing sparse training approaches. The methods are both dense-to-sparse, global magnitude based pruning algorithms, which gradually introduce sparsity during training, utilizing variants of the STE. Specifically, ST-3 adopts soft thresholding and a weight rescaling technique similar to the one used by dropout \cite{srivastava2014dropout}, while Spartan computes a soft top-k mask by solving a regularized Optimal Transportation problem, therefore is more computationally expensive than our approach.

In particular, Table \ref{tab:results_cifar} provides the results obtained on CIFAR-100 using the aforementioned methods over three different architectures and four different levels of sparsity. The reported results correspond to averages of 3 runs with the corresponding standard deviations. The results demonstrate that sparse training with Feather yields steadily more accurate models compared to both SoA methods, either using the simple global pruning approach or combined with ASL\texttt{+}. Notably, the gap in accuracy between our approach and that of the next best performing baseline grows up to 4\% when considering the 99\% sparse ResNet-20 and DenseNet40-24 models. Another interesting remark is that Feather can result to 90\% sparse ResNet-20 and MobileNetV1 models with slightly better generalization accuracies that those of their dense  counterparts, trained using the same number of epochs. The last remark hints that a well designed sparse training method can even be beneficial, not only for producing compact models, but also for improving the generalization performances, when considering relatively conservative sparsity ratios. Note that similar SoA results are observed using both tested backbones, a point that further validates Feather's versatility.

\begin{table}[H]
\centering
\begin{tabular}{lcccc}
\toprule
Ratio & 90\% & 95\% & 98\% & 99\%   \\
\midrule \midrule
\multicolumn{5}{c}{ResNet-20 (1.096M Params):      \res{73.59}{0.44} }   \\ \midrule
ST-3     \cite{vanderschueren2023straight}      &  \res{72.81}{0.13}  & \res{71.72}{0.20}  & \res{67.53}{0.53}  &  \res{58.32}{0.17}  \\
Spartan  \cite{tai2022spartan} 		            &  \res{72.56}{0.35}  & \res{71.60}{0.40}  & \res{67.27}{0.31}  & \res{61.70}{0.21}   \\ 
\hdashline
Feather-Global                                  &  \bres{73.74}{0.17}  & \bres{72.53}{0.32} & \bres{69.83}{0.14} & \bres{65.55}{0.25} \\ 
Feather-ASL\texttt{+}                           &  \res{72.86}{0.10}   & \res{72.42}{0.17}  & \res{69.76}{0.09}  & \res{64.95}{0.47}  \\  \midrule
\multicolumn{5}{c}{ MobileNetV1 (3.315M Params):   \res{71.15}{0.17} }   \\ \midrule
ST-3      \cite{vanderschueren2023straight}		&  \res{70.94}{0.25}  & \res{70.44}{0.23}  & \res{69.40}{0.06}  &  \res{66.63}{0.15} \\
Spartan    \cite{tai2022spartan}	            &  \res{70.52}{0.51}  & \res{69.01}{0.11}  & \res{65.52}{0.24}  &  \res{60.65}{0.22} \\ 
\hdashline
Feather-Global                                  &  \bres{71.55}{0.30} & \res{71.03}{0.20}  & \bres{69.44}{0.29} & \res{67.64}{0.45}  \\
Feather-ASL\texttt{+}                           &  \res{71.10}{0.31}  & \bres{71.26}{0.10} & \res{69.42}{0.12}  & \bres{67.86}{0.03} \\  \midrule
\multicolumn{5}{c}{ DenseNet40-24 (0.714M  Params):   \res{74.70}{0.51} }   \\ \midrule
ST-3     	 \cite{vanderschueren2023straight}	&  \res{72.56}{0.31}  & \res{71.21}{0.35}  & \res{65.48}{0.18}  &  \res{56.18}{0.60} \\
Spartan  	 \cite{tai2022spartan}	            &  \res{73.13}{0.25}  & \res{71.61}{0.04}  &  \res{65.94}{0.07} & \res{58.64}{0.18}  \\ 
\hdashline
Feather-Global                                  &  \res{73.75}{0.36}  & \res{72.36}{0.21}  & \res{69.06}{0.23}  & \bres{63.40}{0.44} \\ 
Feather-ASL\texttt{+}                           &  \bres{73.92}{0.19} & \bres{72.47}{0.12} & \bres{69.08}{0.19} & \res{62.94}{0.14}  \\ 
\bottomrule
\end{tabular}
\vspace{10pt}
\caption{Comparison of Top-1 accuracy on CIFAR-100.}
\label{tab:results_cifar}
\end{table}

\begin{table}
\centering
\begin{tabular}{lcccc}
\toprule
Ratio & 90\% & 95\% & 98\% & 99\%  \\
\midrule\midrule
\multicolumn{5}{c}{ResNet-50 (25.6M Params): $77.10$}  \\ \midrule
GMP      \cite{zhu2017prune} 		        & 73.91  & 70.59  & 57.90 & 44.78 \\
DNW		 \cite{wortsman2019discovering}	    & 74.00  & 68.30  & 58.20 & - 	  \\
STR      \cite{kusupati2020soft} 		    & 74.31  & 70.40  & 61.46 & 51.82 \\  
ProbMask \cite{zhou2021effective}  		    & 74.68  & 71.50  & 66.83 & 61.07 \\
OptG	 \cite{zhang2022optimizing}	        & 74.28  & 72.45  & 67.20 & 62.10 \\
ST-3     \cite{vanderschueren2023straight}  & 76.03  & 74.46  & 70.46 & 63.88 \\
Spartan  \cite{tai2022spartan} 		        & 76.17  & 74.68  &   -   & 63.87 \\ 
\hdashline
Feather-Global      & \textbf{76.93} & \textbf{75.27}   & \textbf{72.92}  & \textbf{68.85}   \\
\bottomrule
\end{tabular}
\vspace{10pt}
\caption{Comparison of Top-1 accuracy on ImageNet.}
\label{tab:results_imagenet}
\end{table}

\noindent \textbf{ImageNet:} 
\noindent For the ImageNet experiments we include results from literature from an extended number of pruning methods achieved using the same number of epochs (100) and data augmentation. In particular, results from the relevant methods GMP \cite{zhu2017prune}, DNW \cite{wortsman2019discovering}, STR \cite{kusupati2020soft}, ProbMask \cite{zhou2021effective},  OptG \cite{zhang2022optimizing}, ST-3 \cite{vanderschueren2023straight} and Spartan \cite{tai2022spartan} are presented in Table \ref{tab:results_imagenet}. We adopt the global pruning scheme combined with Feather module, being conceptually the simplest approach, in order to highlight our method's effectiveness compared to more sophisticated baselines. As we can see, it provides considerably better results than those from previous SoA, especially at very challenging pruning ratios over 98\%. We want to emphasize that the improved performance with Feather does not come at the cost of higher training overheads or the need for complicated hyperparameter settings, in contrast to certain baselines (\eg \cite{kusupati2020soft, zhou2021effective}). Instead, the resulting accuracy gains can be attributed to the simple, yet carefully formulated modifications of the proposed module that fully utilizes the STE-based sparse training approach.

\section{Conclusions }

This paper proposes Feather, an effective and efficient sparse training module that can be easily applied to pruning frameworks. In particular, as demonstrated by extensive experiments on CIFAR and ImageNet datasets, using both a global and a layer-wise approach, it results to improving the previous SoA results, especially at high pruning ratios. Furthermore, ours method's success indicates the large potential of properly understanding and consequently improving the sparse training dynamics using an STE based approach, that despite its simplicity is shown to be highly effective.

\section*{Acknowledgements}

This research work was supported by the Hellenic Foundation for Research \& Innovation (H.F.R.I.) under the "2nd Call for H.F.R.I. Research Projects to support Faculty Members \& Researchers" (Project Number:2656, Acronym: TROGEMAL).

\bibliographystyle{unsrt}
\bibliography{references}  

\vfill

\pagebreak

\appendix

\section{Appendix}

\subsection{Training Hyperparameters}

\begin{table}[h]
\begin{center}
\begin{tabular}{lcc}
\toprule
Dataset 	& CIFAR-100 & ImageNet  \\
\midrule\midrule
Epochs 		 & 	160		& 	100		\\
Batch Size   & 	128		& 	256		\\
Weight Decay & 5e-4     &  5e-5    \\
Optimizer    &   SGD    &    SGD		\\
LR			 & 0.1      &   0.2		\\
LR-Scheduler & Cosine   & Cosine+Warmup\\
Momentum 	 & 0.9	    &  0.9		\\
Label Smoothing & -    & 0.1\\ 
\bottomrule
\end{tabular}
\end{center}
\vspace{10pt}
\caption{ Training hyperparameters used for all our experiments on CIFAR-100 and ImageNet datasets.}
\label{tab:hyper_feather}
\end{table}

Table \ref{tab:hyper_feather} summarizes the training hyperparameters used for our experiments on CIFAR-100 \cite{krizhevsky2009learning} and ImageNet \cite{russakovsky2015imagenet} datasets. The chosen hyperparameters  are selected based on standard practices for the particular datasets and are kept the same regardless the network architecture or the target sparsity ratio  (in contrast to \eg \cite{kusupati2020soft,  vanderschueren2023straight} where the Weight Decay is adjusted among different runs, based on the target sparsity ratio).  By adopting commonly used hyperparameters and keeping them unchanged among all our experiments we opted to show that our method is able to achieve SoA results  without the need of fine-tuning and complicated training configurations.

\subsection{Impact of Threshold's p-value}

\begin{figure}[h]
\begin{center}
\subfloat[]{\includegraphics[width=.45\linewidth]{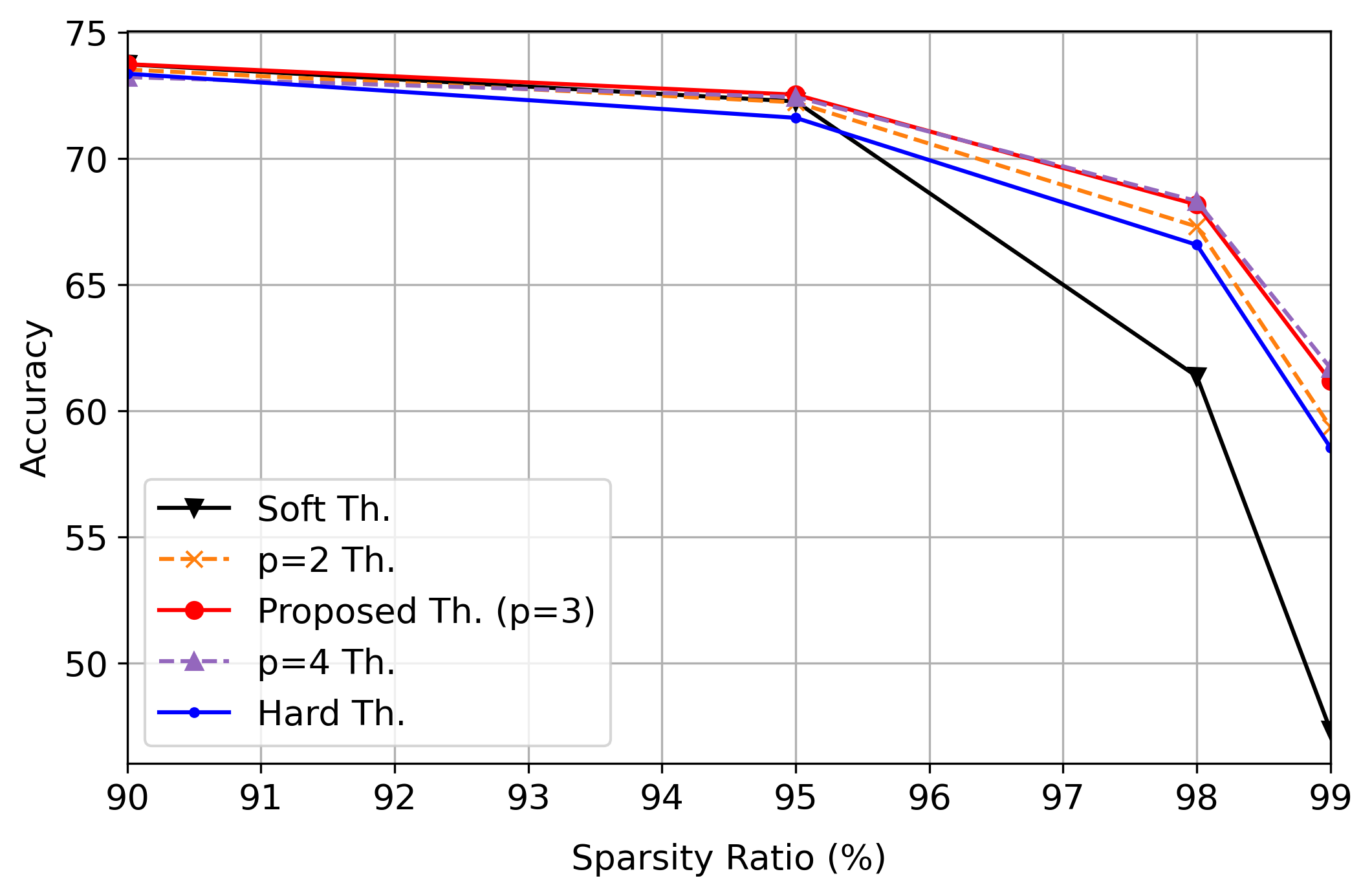}
\label{fig:abl4_a}
}
\subfloat[]{\includegraphics[width=.41\linewidth]{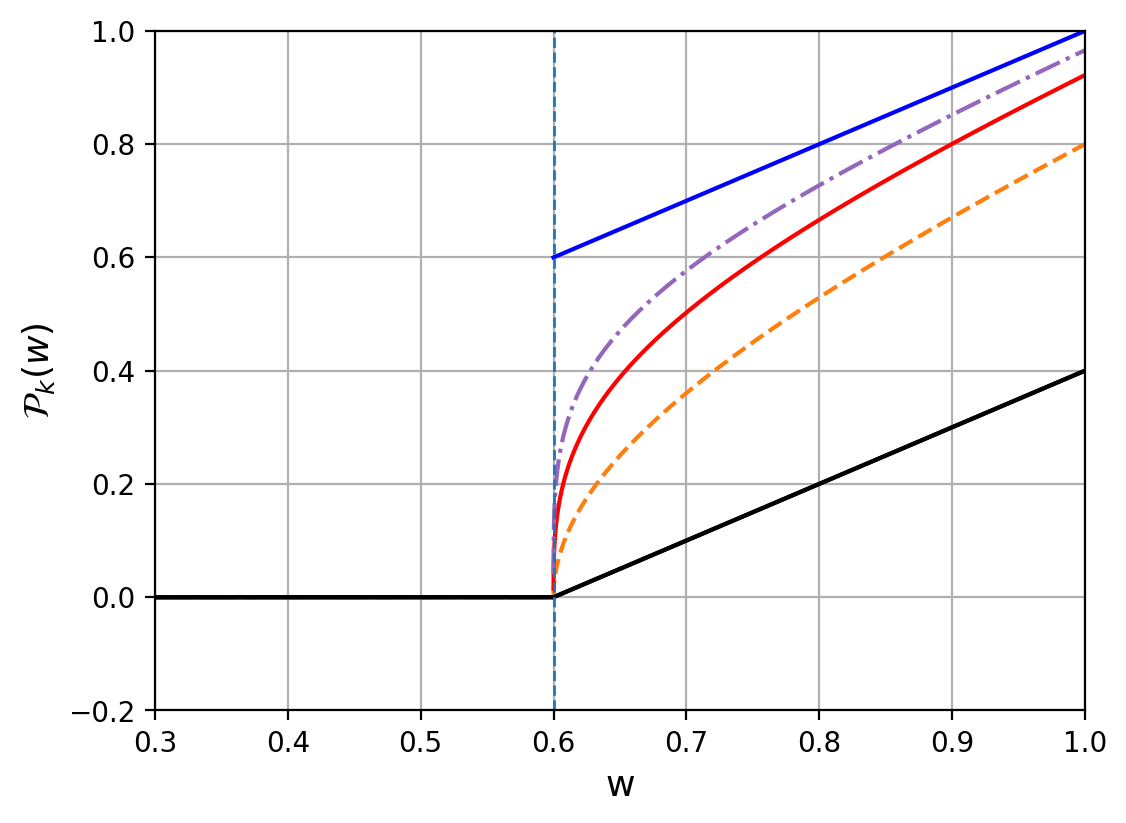}
\label{fig:abl4_b}
}
\caption{A study of the effect of the $p$ value of the proposed family of thresholds on the final sparse model accuracy. Results from ResNet-20 trained on CIFAR-100 (a) and the corresponding thresholds used (b).}
\label{fig:abl4}
\end{center}
\end{figure}

The proposed threshold with $p=3$ is compared with the ones with $p=2$ and $p=4$ in Figure \ref{fig:abl4}. We observe that $p=3$ is preferable to $p=2$ based on the resulting final accuracy while  $p=4$ results to no further improvement (Figure \ref{fig:abl4_a}). Due to that, $p=3$ is chosen to
give a fine balanced threshold between the two extremes, Hard and Soft thresholds respectively, although, as shown, good results are obtainable even with values of $p$ near 3. A reasonable explanation for the slight under-performance using $p=2$ is that the resulting threshold still leads to a considerable amount of shrinkage (Figure \ref{fig:abl4_b}), thus induces more bias between the thresholded weights and their dense counterparts. Notably, even for $p=2$ the results are favorable compared to those obtained by using the Hard and Soft Thresholds, further validating the robustness of our family of threshold operators.

\subsection{Stability of the Sparsity Mask \emph{vs}.\ Gradient Scaling}

\begin{figure}[ht]
	\centering
  \includegraphics[width=.6\linewidth]{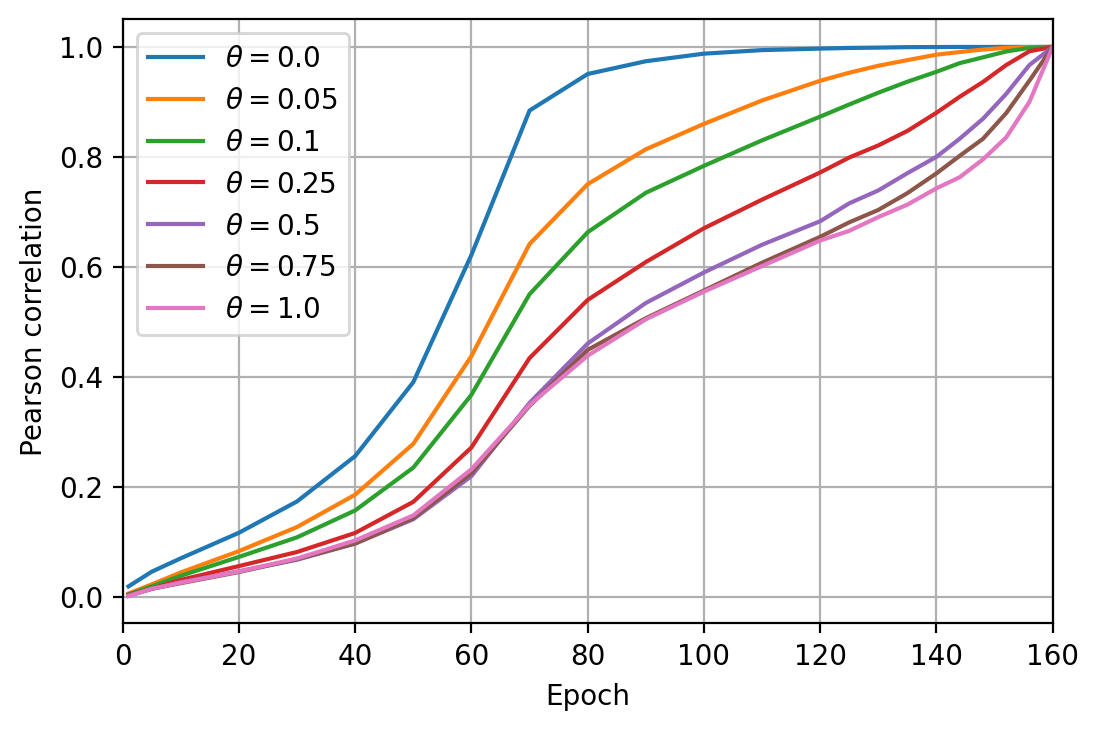}
\caption{ Plot of Pearson correlation coefficients between the sparsity mask obtained at the end of each epoch and the mask at the end of training, for varying values of the gradient scaling parameter $\theta$. Results from ResNet-20 trained on CIFAR-100.} 
\label{fig:abl5}
\end{figure}

Figure \ref{fig:abl5} empirically validates that the gradient scaling parameter $\theta \in [0,1]$ influences the stability of the sparsity mask, \ie the mask that indicates which parameters are pruned and which are active during sparse training.  Specifically, for each experiment, using a specified (constant) value for $\theta$, the Pearson correlation coefficients between the mask at the end of every training epoch and the final mask, obtained at the end of training, are shown. Experiments with $\theta$ near zero result to curves that converge to 1 more rapidly, compared to the ones from experiments with $\theta$ close to unity. This indicates that when using $\theta$ near zero (or at the extreme case $\theta=0$) the sparsity mask (and thus the sparsity pattern) is stabilized earlier during the training process, compared to when using larger values of $\theta$. Based on our empirical analysis, the suitable amount of stability for the sparsity mask relates to the sparsity target; The higher the requested final sparsity the more beneficial is to keep the mask more stable (up to a reasonable extent) to avoid destabilizing the highly pruned network.  We note that the mask's stability is also studied in \cite{tai2022spartan}, where a soft top-k mask is computed by solving a regularized Optimal Transportation problem in order to regulate its stability, although our approach using gradient scaling (combined with the proposed threshold operator) is considerably less computationally expensive while resulting to favorable final accuracies.

\subsection{Feather Improves Pruning Backbones}

\begin{figure}[H]
\begin{center}
\subfloat[]{\includegraphics[width=.45\linewidth]{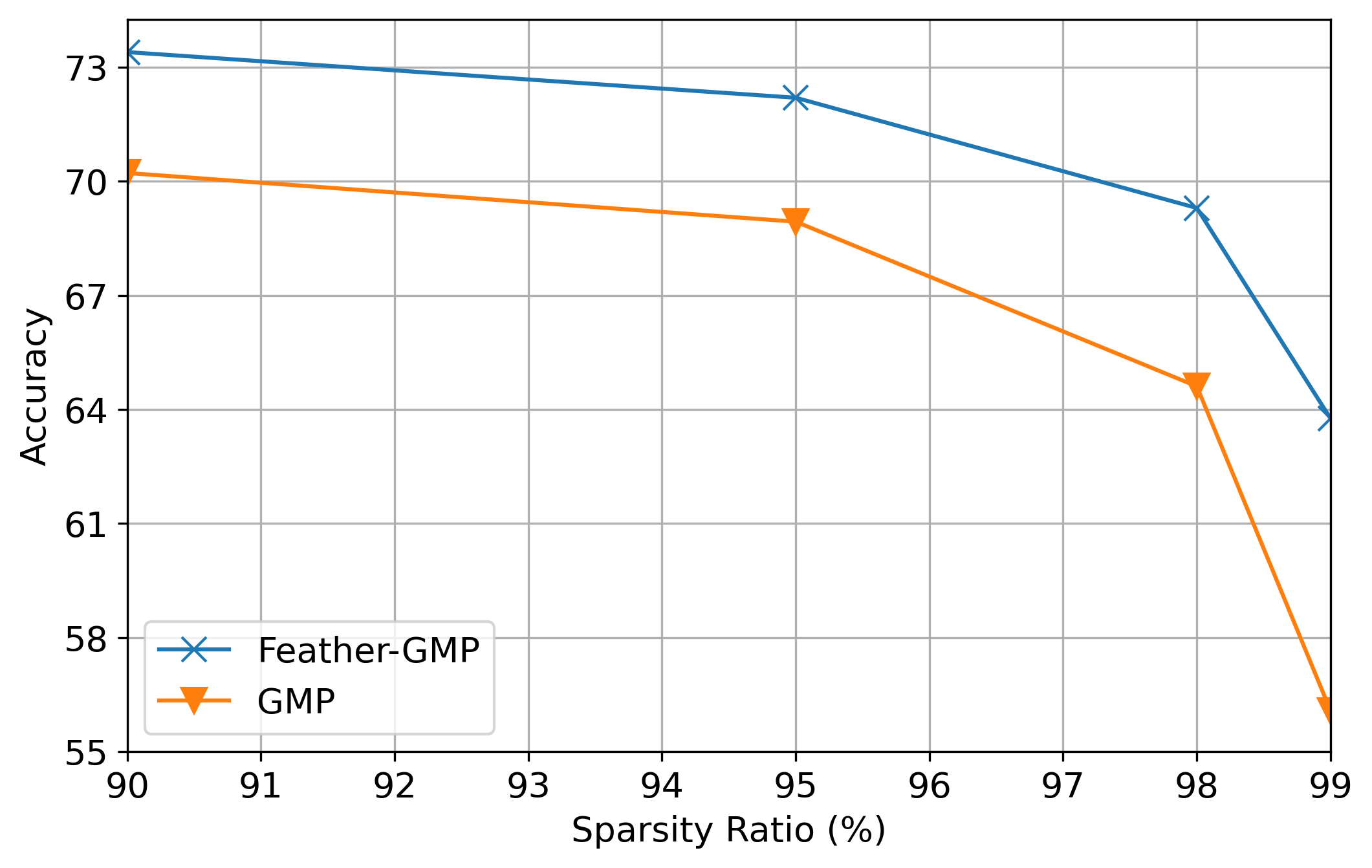}
\label{fig:abl6}
}
\subfloat[]{\includegraphics[width=.45\linewidth]{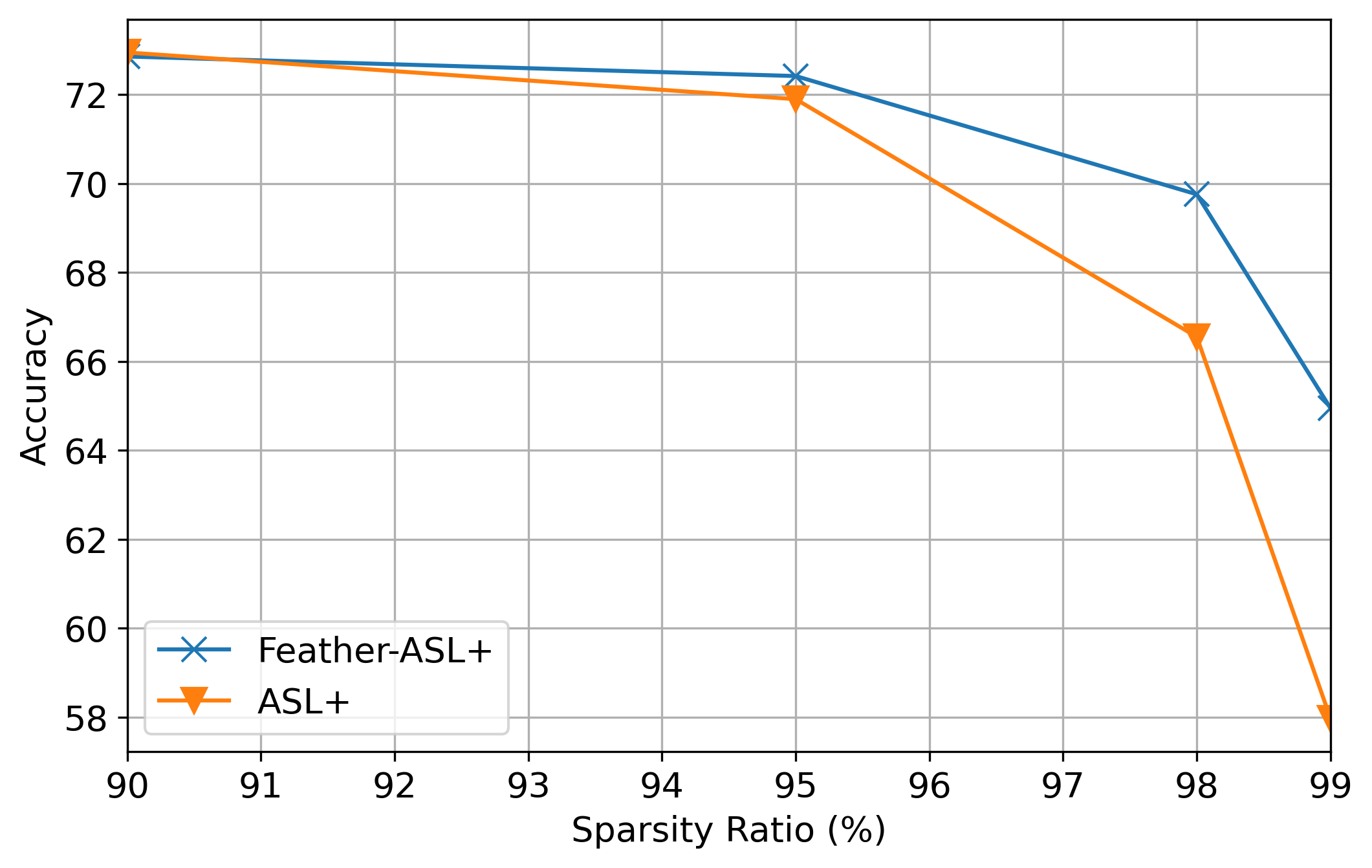}
\label{fig:abl7}
}
\caption{ Feather improves the accuracy of common sparse training backbones: (a) GMP, a uniform layer-wise sparsity pruning backbone (b) the ASL\texttt{+} framework. Results from ResNet-20 trained on CIFAR-100.}
\label{fig:abl67}
\end{center}
\end{figure}

Combining the Feather module with existing backbones results to more accurate networks, as shown in Figure \ref{fig:abl67}. In \ref{fig:abl6} Feather is used to improve the accuracy of GMP \cite{zhu2017prune}, a layer-wise magnitude pruning backbone that prunes all layers\footnote{With the exception of the first convolutional layer, which was left dense when using GMP in our experiments due to having a very small number of parameters.}
to the same (uniform) amount of sparsity, gradually increasing the pruning ratio. Our module significantly improves the resulting accuracy when combined with the very simplistic GMP backbone. Furthermore, in \ref{fig:abl7} we compare the accuracy of the sparse models obtained with Feather combined with ASL\texttt{+} \cite{retsinas2021online} and the ones using only ASL\texttt{+}, showing that our module leads to accuracy improvements for the challenging sparsity ratios (95\% and above).

\subsection{MobileNetV1 on ImageNet}

\begin{table}[H]
\begin{center}
\begin{tabular}{lcc}
\toprule
Ratio & 89\% & 94.1\%   \\
\midrule\midrule
\multicolumn{3}{c}{MobileNetV1 ( 4.21M Params): $71.95$}  \\ \midrule
GMP \cite{zhu2017prune}       		 	 & 61.80  &   -			\\
STR \cite{kusupati2020soft}     		 & 62.10  &   - 		\\ 
ProbMask \cite{zhou2021effective} 		 & 65.19  & 60.10    	\\
ST-3 \cite{vanderschueren2023straight}   & 66.67  &  61.19  	\\
\hdashline
Feather-Global      					& \textbf{68.13} & \textbf{63.63}  \\
\bottomrule
\end{tabular}
\end{center}
\vspace{10pt}
\caption{Top-1 accuracy of MobileNetV1 on ImageNet.}
\label{tab:mobnet_imagenet_results}
\end{table}

In Table \ref{tab:mobnet_imagenet_results} we provide additional experiments on ImageNet \cite{russakovsky2015imagenet} using the MobileNetV1 \cite{howard2017mobilenets} architecture. More specifically, we compare the accuracies obtained by using Feather combined with the global pruning backbone with the ones from GMP \cite{zhu2017prune}, STR \cite{kusupati2020soft}, ProbMask \cite{zhou2021effective} and  ST-3 \cite{vanderschueren2023straight} which report results for the 89\% and 94.1\% sparsity ratios, using the same number of epochs (100) and data augmentation as in our experiments. Our approach surpasses the previous SoA by 1.46\% and 2.44\% Top-1 accuracy at the 89\% and 94.1\% sparsity ratios respectively, a result that further validates Feather's  effectiveness and generalization abilities on large datasets with different model architectures.

\subsection{Accuracy \emph{vs}.\ FLOP Measurements}

\begin{figure}[H]
	\centering
  \includegraphics[width=.5\linewidth]{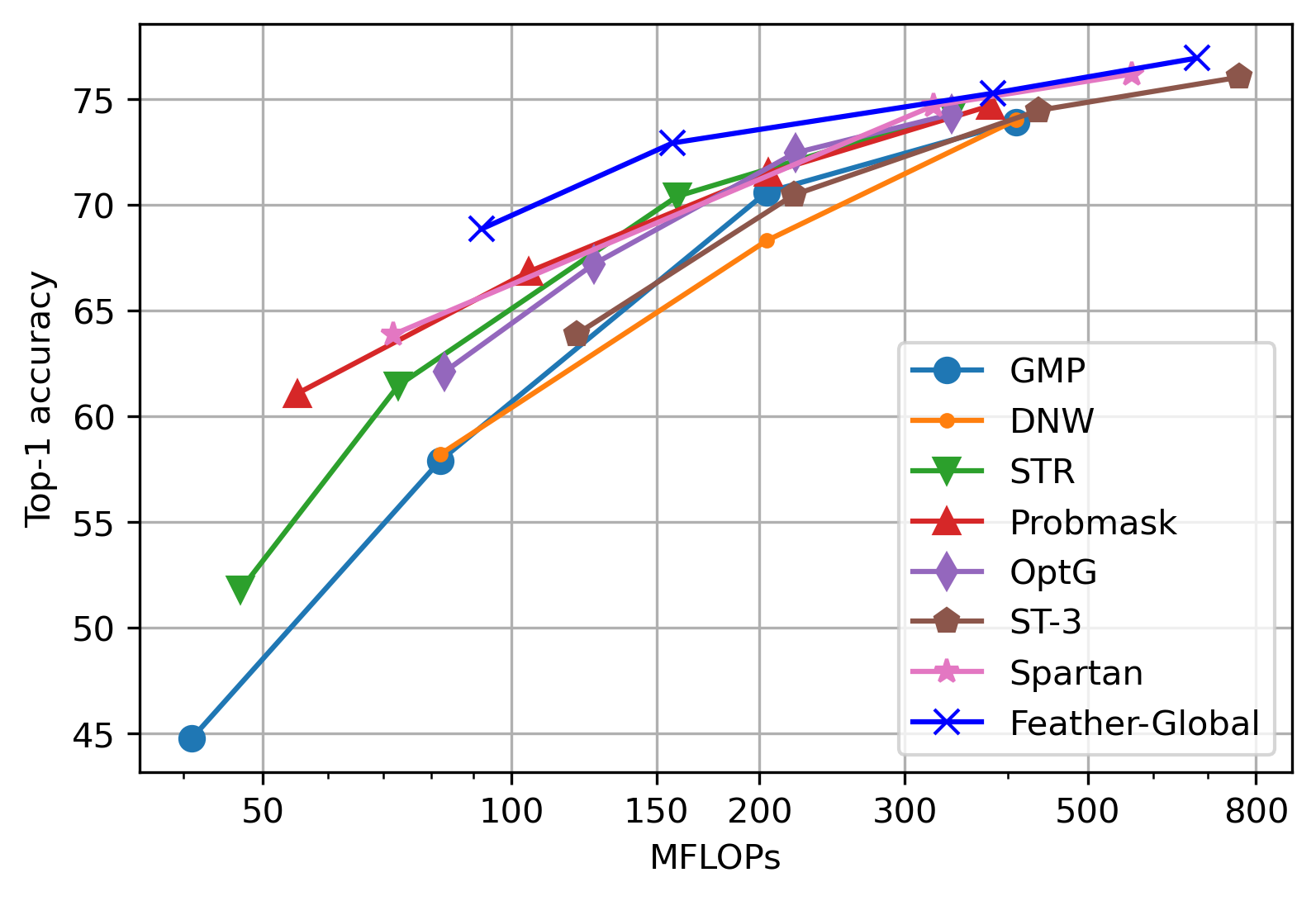}
\caption{ Top-1 accuracy \vs FLOPs of ResNet-50 on ImageNet.} 
\label{fig:acc_vs_flops}
\end{figure}

The Feather module, combined with the global pruning backbone, leads to favorable Top-1 accuracy results over the ones from the baselines under similar FLOPs requirements of the sparsified ResNet-50 \cite{he2016deep}, as shown by the frontier curve in Figure \ref{fig:acc_vs_flops}. We note that the per-layer sparsity distribution obtained by the global pruning backbone by default does not prioritize FLOPs reduction, while layer-wise methods such as GMP \cite{zhu2017prune} and STR \cite{kusupati2020soft} tend to result to sparse models with minimum FLOPs for a given sparsity ratio, although at a cost of considerable accuracy drops.

While extended analysis on optimizing FLOPs for a given sparsity target is not the scope of this work, to further showcase the efficacy of Feather we experimented with biasing the global pruning backbone towards pruning earlier layers more aggressively, as suggested in \cite{vanderschueren2023straight}. With the FLOPs-biased global pruning backbone, training the ResNet-50 on ImageNet at 99\% sparsity, Feather resulted in a model with \textbf{67.2}\% \textbf{Top-1 accuracy}, now requiring only \textbf{42MFLOPs}. Therefore, the superior accuracy of our sparse model was greatly preserved,  still achieving the best accuracy (by a 3.3\% margin) among the baselines at the 99\% ratio, now for considerably fewer FLOP requirements, matching those of  GMP (41MFLOPs), the baseline resulting to the fewer FLOPs, although having accuracy more than 20\% higher. Having showcased Feather's great potential at obtaining models with superior accuracy and FLOPs, we leave further experimentation (possibly with more sophisticated backbones) as future work.

\end{document}